\documentclass[runningheads]{llncs}
\usepackage{graphicx}
\usepackage{amsfonts}
\usepackage{amsmath}
\usepackage{subfigure}
\usepackage{makecell}
\begin{document}
\title{CILF:Causality Inspired Learning Framework for Out-of-Distribution Vehicle Trajectory Prediction}

%
%
\author{Shengyi Li\and
Qifan Xue\and
Yezhuo Zhang\and 
Xuanpeng Li\thanks{Corresponding Author}}
\authorrunning{S. Li et al.}
%
\institute{School of Instrument Science and Engineering, Southeast University, Nanjing 211189, China\\
\email{li\_shengyi@seu.edu.cn\\xue\_qifan@seu.edu.cn\\zhang\_yezhuo@seu.edu.cn\\li\_xuanpeng@seu.edu.cn}}
\maketitle              
\begin{abstract}
Trajectory prediction is critical for autonomous driving vehicles.
Most existing methods tend to model the correlation between history trajectory (input) and future trajectory (output).
Since correlation is just a superficial description of reality, these methods rely heavily on the i.i.d. assumption
and evince a heightened susceptibility to out-of-distribution data.
To address this problem,
we propose an \textbf{O}ut-\textbf{o}f-\textbf{D}istribution \textbf{C}ausal \textbf{G}raph (OOD-CG),
which explicitly defines the underlying causal structure of the data with three entangled latent features:
1) domain-invariant causal feature (\textit{IC}),
2) domain-variant causal feature (\textit{VC}), and
3) domain-variant non-causal feature (\textit{VN}).
While these features are confounded by confounder (\textit{C}) and domain selector (\textit{D}).
To leverage  causal features for prediction,
we propose a \textbf{C}ausal \textbf{I}nspired \textbf{L}earning \textbf{F}ramework (CILF),
which includes three steps:
1) extracting domain-invariant causal feature by means of an invariance loss,
2) extracting domain variant feature by domain contrastive learning, and 
3) separating domain-variant causal and non-causal feature by encouraging causal sufficiency.
We evaluate the performance of CILF
in different vehicle trajectory prediction models
on the mainstream datasets NGSIM and INTERACTION.
Experiments show promising improvements in CILF on domain generalization.
\end{abstract}
\section{Introduction}
Trajectory prediction is essential for both the perception and planning modules of autonomous vehicles~\cite{paden2016survey,lefevre2014survey}
in order to reduce the risk of collisions~\cite{li2022recursive}.
Recent trajectory prediction methods are primarily built with deep neural networks,
which are trained to model the correlation between history trajectory and future trajectory.
The robustness of such correlation is guaranteed by the independent and identically distributed (i.i.d.) assumption.
As a result, the model trained on i.i.d. samples often fails to be generalized to out-of-distribution (OOD) samples.

Recently, there has been a growing interest in utilizing causal representation learning~\cite{scholkopf2021toward}
to tackle the challenge of out-of-domain generalization~\cite{zhou2022domain}.
Causal representation learning is based on the Structure Causal Model (SCM)~\cite{glymour2016causal,pearl2009causality},
a mathematical tool for modeling human metaphysical concepts about causation.
Causal representation learning enables the model to discern the underlying causal structure
of data by incorporating causal-related prior knowledge into the model.

\begin{figure}
  \centering
  \subfigure[]{
    \includegraphics[width=0.22\linewidth]{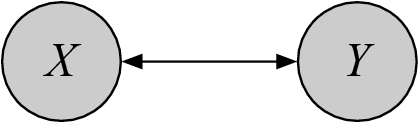}
    \label{fig1-1}
  }
  \subfigure[]{
    \includegraphics[width=0.22\linewidth]{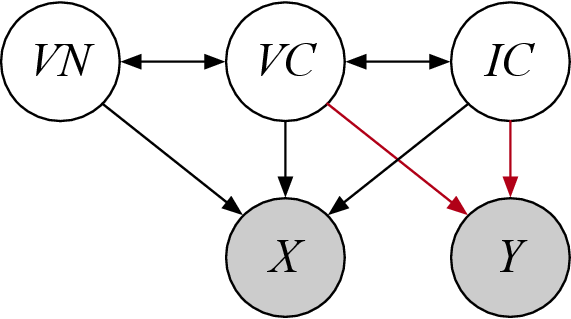}
    \label{fig1-2}
  }
  \subfigure[]{
    \includegraphics[width=0.22\linewidth]{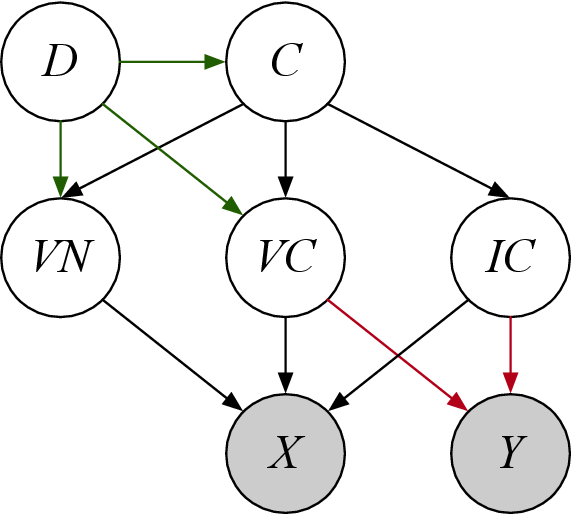}
    \label{fig1-3}
  }
  \subfigure[]{
    \includegraphics[width=0.22\linewidth]{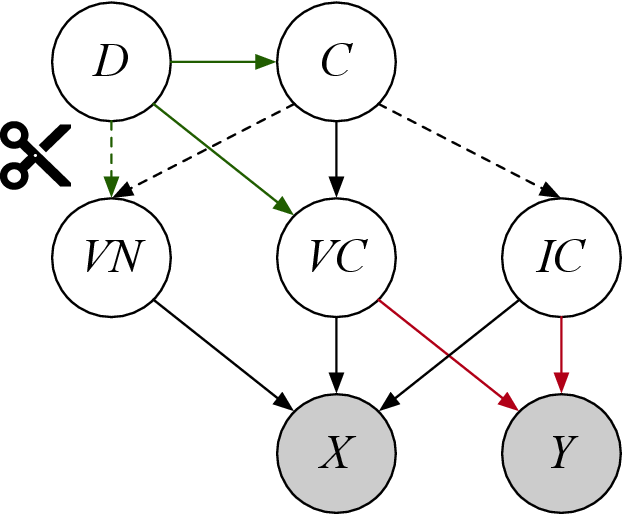}
    \label{fig1-4}
  }
  \caption{
      The step-by-step introduction of OOD-CG.
      White nodes represent latent features;
      gray nodes represent observable variables;
      bidirectional arrows represent correlations;
      unidirectional arrows represent causal mechanisms;
      red arrows represent causal mechanisms critical to prediction;
      and green arrows represent domain effects.
  }\label{fig1}
\end{figure}

This paper proposes an Out-of-Distribution Causal Graph (OOD-CG) based on SCM, as shown in Fig.\ref{fig1}.
OOD-DG divides the latent features into three categories:
1) Domain-invariant Causal Feature (\textit{IC}) such as physical laws, driving habits,  etc.
2) Domain-variant Causal Feature (\textit{VC}) such as road traffic flow, traffic scenes,  etc.
3) Domain-variant Non-causal Feature (\textit{VN}) like sensor noise,  etc.
These features are entangled due to the confounding effects of
backdoor confounder (\textit{C}) and domain selector (\textit{D}).

To leverage  causal features for trajectory prediction,
we introduce a Causal-Inspired Learning Framework (CILF)
based on causal representation learning.
CILF includes three parts to block the backdoor paths 
associated with \textit{IC}, \textit{VC}, and \textit{VN}.
First, to block the backdoor path between \textit{IC} and domain-variant feature \textit{V},
CILF utilizes invariant risk minimization (IRM)~\cite{arjovsky2019invariant} to extract \textit{IC}
and domain contrastive learning~\cite{khosla2020supervised} to extract \textit{V}.
Then, to block the backdoor path between \textit{VC} and \textit{VN},
CILF introduces domain adversarial learning~\cite{jang2016categorical} to separate \textit{VC} and \textit{VN}.

Our contributions can be summarized as follows:

(1) We propose a theoretical model named OOD-CG, which explicitly elucidates
the causal mechanisms and causal structure of the out-of-distribution generalization problem.

(2) Based on causal representation learning, we propose a learning framework called CILF 
for out-of-distribution vehicle trajectory prediction.
CILF contains three steps to block the backdoor connection associated with \textit{IC}, \textit{VC}, and \textit{VN},
allowing the model to employ causal features for prediction.

\section{Related Work}
\subsubsection{Vehicle trajectory Prediction}
Recent works widely employ the sequence-to-sequence (Seq2seq) framework~\cite{sutskever2014sequence}
to predict a vehicle's future trajectory based on its history trajectory~\cite{alahi2016social,deo2018convolutional,lee2017desire,tang2019multiple}.
Alahi et al. introduce S-LSTM~\cite{alahi2016social}, which incorporates a social pooling mechanism to aggregate and encode the social behaviors of surrounding vehicles.
Deo et al. propose CS-LSTM~\cite{deo2018convolutional}, which utilizes convolutional operations to enhance the model's performance.
Lee et al. introduce a trajectory prediction model called DESIRE~\cite{lee2017desire}, which combines conditional variational auto-encoder (CVAE) and GRU to generate multimodal predictions of future trajectories.
Tang et al. incorporate an attention mechanism into the Seq2seq framework named MFP~\cite{tang2019multiple},
a model that can effectively learn motion representations of multiple vehicles across multiple time steps.
However, current vehicle trajectory prediction approaches still face the challenge
of OOD generalization, posing a serious threat to the safety of autonomous vehicles.

\subsubsection{Out-of-Distribution generalization} 
Previous methods handle OOD generalization in two paradigms:
domain adaptation (DA) and domain generalization (DG)~\cite{zhou2022domain}.
DA allows the model to access a small portion of unlabeled target domain data
during training, thereby reducing the difficulty of the OOD problem to some extent.
DA aims to learn an embedding space
where source domain samples and target domain samples follow similar distributions
via minimizing divergence~\cite{gretton2012kernel,long2017deep},
domain adversarial learning~\cite{ganin2016domain,tzeng2017adversarial}, etc.
DG prohibits models from accessing any form of target domain data during training.
DG aims to learn knowledge that can be directly transferred to unknown target domains.
Relevant literature has proposed a range of solutions, including contrastive learning~\cite{motiian2017unified,yoon2019generalizable},
domain adversarial learning~\cite{tzeng2017adversarial,zhang2018collaborative}, etc.
As for vehicle trajectory prediction, it is difficult
to acquire even unlabeled target domain samples. As a result, we follow the paradigm of domain generalization.
\subsubsection{Causality inspired approaches for domain generalization} 
There has been a growing trend toward utilizing causality inspired methods to address the OOD problem.
Some of them suggest using mathematical formulas from causal inference~\cite{pearl2009causality},
e.g.,front-door adjustment~\cite{nguyen2022front} and back-door adjustment~\cite{bagi2023generative} formulas,
to directly derive the loss function.
However, these methods often make strong assumptions (e.g.,restricting the model to be a variational auto-encoder~\cite{bagi2023generative}),
which decrease the diversity of the model's hypothesis space and
limit the applicability of the model.
Some approaches instead only make assumptions about the general causal structure of the OOD problem
in order to guide the design of network structures~\cite{liu2022towards,lv2022causality}.
We follow this approach in order to enhance the versatility of the proposed model.

\section{Theoretical Analysis}
\subsection{Problem Formulation}
We formulated vehicle trajectory prediction as estimating future trajectories
$Y_i=\{(x_{i,t},y_{i,t}) \in \mathbb{R}^2|t=t_{obs}+1,...,t_{pred}\}$ 
based on the observed history trajectories
$X_i=\{(x_{i,t},y_{i,t}) \in \mathbb{R}^2|t=1,2,...,t_{obs}\}$
of $N$ visible vehicles in the current scene.
In the context of domain generalization, the training dataset is collected from $K$ source domains,
denoted as $S\in\{S_1,S_2,...,S_K\}$, while the test dataset is collected from $M$ target domains, denoted
as $T\in\{T_1,T_2,...,T_M\}$.
\subsection{OOD-CG}
\textbf{Fig.\ref{fig1-1}}.
Traditional deep learning is designed
to capture statistical correlations between inputs $X$ and outputs $Y$. 
Correlations are obviously inadequate to solve the OOD problem.
Fortunately, Reichenbach provides us with a strong tool called the common causal principle~\cite{reichenbach1956direction}
to decompose these correlations into a set of backdoor features.
\begin{theorem}
  Common causal principle: if two random variables $X$ and $Y$ are correlated,
  there must exist another random variable $S$ that has causal relationships with
  both $X$ and $Y$. Furthermore, $S$ can completely substitute for the correlations
  between $X$ and $Y$, i.e., $X \perp\!\!\!\perp Y | S$.
  \label{theorem-1}
\end{theorem}

\textbf{Fig.\ref{fig1-2}}. 
We divide these backdoor features into three classes:
(1) Domain-Invariant Causal Feature (\textit{IC}): Driving habits, physical laws,  etc.
(2) Domain-Variant Causal Feature (\textit{VC}): Traffic density, traffic scenario,  etc.
(3) Domain-Variant Non-causal Feature (\textit{VN}): Sensor measurement noise,  etc.
Bidirectional arrows are used in this step to represent the entanglement between these backdoor features.
Previous studies often ignore domain-variant causal feature \textit{VC},
and focus solely on utilizing domain-invariant causal feature \textit{IC}
for prediction~\cite{bagi2023generative,chen2021human,nguyen2022front}.
These approaches utilize insufficient causal information for prediction and thus fail to
simultaneously improve prediction accuracy in both the source and target domains.

\textbf{Fig.\ref{fig1-3}}.
According to Theorem.\ref{theorem-1},
we introduce a Confounder (\textit{C})
to summarize all the correlations among \textit{IC}, \textit{VC}, and \textit{VN}.
We also introduce a Domain Selector (\textit{D}) to represent domain effects.
Since domain label is unavailable during testing,
we treat \textit{D} as an unobservable latent variable consistent with ~\cite{bagi2023generative,lv2022causality}.
Now we can concretize distribution shifts as differences
in feature prior distributions $P(C),P(VC),P(VN)$ and
causal mechanism conditional distributions $P(VC|C),P(VN|C)$
between source and target domains.

\textbf{Fig.\ref{fig1-4}}.
In order to extract domain-invariant causal mechanisms $P(Y|VC)$, $P(Y|IC)$
for prediction, it is necessary to block the backdoor paths connecting these entangled features.
To do so, we only need to block the backdoor paths of \textit{IC} and \textit{VN},
which are represented by dashed lines in the figure.
Once done, \textit{VC} actually serves as a mediator~\cite{glymour2016causal,pearl2009causality}
in causal effects from \textit{D} and \textit{C} to Y, which can be completely substituted by \textit{VC}.
In the next chapter, we introduce a causal-inspired learning framework (CILF) for
OOD vehicle trajectory prediction, encouraging models to block the backdoor paths of \textit{IC} and \textit{VN}.

\section{CILF}
\begin{figure}
  \centering
  \includegraphics[width=1.0\linewidth]{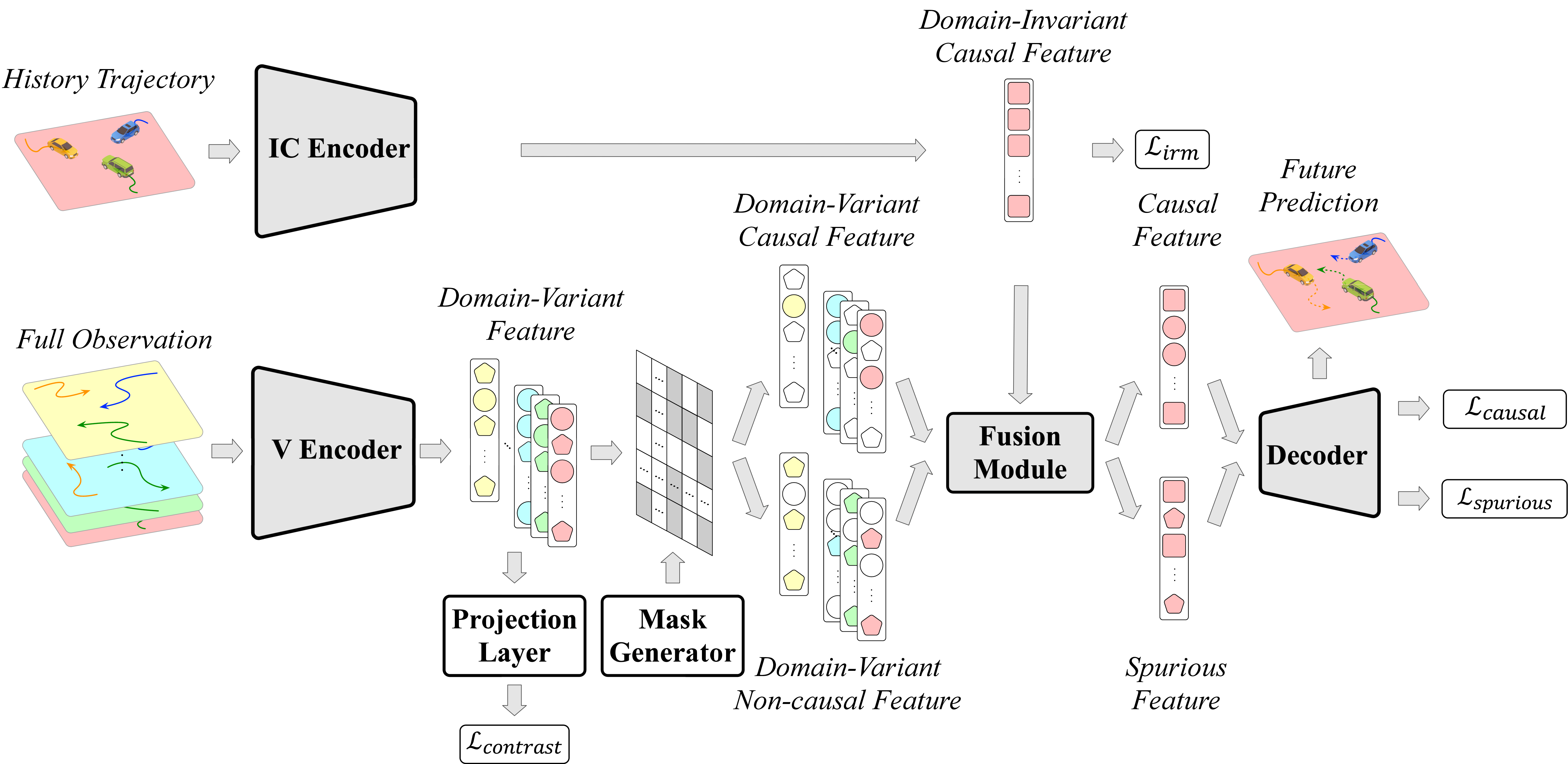}
  \caption{
      CILF consists of three parts:
      (i)\textit{IC} encoder $\Phi_{ic}(.)$ takes history trajectories of the current vehicle as input.
      The IRM loss $\mathcal{L}_{irm}$ is utilized to encourage the extracted feature to exhibit domain invariance;
      (ii)Domain-variant feature encoder $\Phi_{v}(.)$ takes full observations from each source domain
      as input, such as history trajectories of neighboring vehicles, map information,  etc.
      The domain contrastive learning loss $\mathcal{L}_{contrast}$ is utilized to encourage
      the extracted feature to exhibit domain variance;
      (iii)Mask generator $M(.)$ generates a causally sufficient mask to select
      the causal dimensions \textit{VC} and non-causal dimensions \textit{VN} from \textit{V}.
  }\label{fig2}
\end{figure}

Compared to traditional machine learning,
deep learning requires searching within
an extremely complex hypothesis space.
As a result, researchers often introduce inductive bias
into deep learning models to reduce search difficulty,
allowing models to find some relatively acceptable local optimums
within limited training iterations.
OOD-CG is proposed to impose such inductive bias theoretically.
Guided by OOD-CG, we propose CILF (see Fig.\ref{fig2}) to extract causal features for prediction.
First, to block the backdoor connection between \textit{IC} and \textit{V}, CILF introduces two kinds of losses: 
IRM loss to encourage domain-invariant causal feature \textit{IC};
Domain contrastive loss to encourage domain-variant feature \textit{V}.
Then, to block the backdoor path between \textit{VC} and \textit{VN}, CILF introduces
domain adversarial learning to train a mask generator able to detect the causal dimensions \textit{VC} and 
non-causal dimensions \textit{VN} of domain-variant feature \textit{V}.

\subsection{Extract Domain-invariant Causal Feature}
Domain-invariant features are the intersection of feature sets from different domains.
The non-causal information within these features exhibits a significant reduction after
taking the intersection.
As a result, extracting \textit{IC} can be approximated as extracting the domain-invariant feature.
The definition of a domain-invariant encoder is given as follows~\cite{arjovsky2019invariant}:

\begin{definition}
  Domain-invariant feature encoder: An encoder $\Phi(.):\mathcal{X}\to\mathcal{H}$ is said to be
  a domain-invariant encoder across all domains $s\in S$ if and only if there
  exists a decoder $\Theta(.):\mathcal{H}\to\mathcal{Y}$ that achieves optimum across all domains.
  This condition can be further formulated as a conditional optimization problem:
  \begin{equation} 
    \begin{split}\label{eq1}
      &\underset{\Phi,\Theta}{\mathrm{min}}\Sigma_{s\in S}\mathcal{L}_s(\Theta \circ \Phi),\\
      \text{subject to } \Theta \in &\underset{\bar{\Theta}}{\mathrm{argmin}}\mathcal{L}_s(\bar{\Theta}\circ\Phi),\text{for all }s\in S.
    \end{split}
  \end{equation}
\end{definition}

However, this is a bi-level optimization problem, which is computationally difficult to solve.
In current research, the problem is often relaxed to a gradient regularization penalty based on
empirical risk minimization~\cite{arjovsky2019invariant}:
\begin{equation}\label{eq2}
  \underset{\Phi_{IC},\Theta}{\mathrm{min}}
  \frac{1}{\lvert S\rvert}\Sigma_{s\in S}[\mathcal{L}_s(\Theta \circ \Phi_{IC})+
  \lambda \lVert\nabla_{\Theta}\mathcal{L}_s(\Theta \circ \Phi_{IC})\rVert^2],
\end{equation}
\begin{equation}\label{eq3}
  \mathcal{L}_{irm}=\frac{1}{\lvert S\rvert}
  \lVert\nabla_{\Theta}\mathcal{L}_s(\Theta \circ \Phi_{IC})\rVert^2,
\end{equation}
where $\mathcal{L}_{irm}$ denotes the IRM loss,
$\lambda$ denotes the balance parameter.

We separate $\Phi_{IC}$ from the backbone to further block the backdoor path of \textit{IC}, as shown in Fig.\ref{fig2}.
$\Phi_{IC}$ only accepts the history trajectories of the current vehicle as input,
which contains relatively limited domain knowledge.

\subsection{Extract Domain-variant Feature}
To extract domain-variant feature, we design $\Phi_v(.)$, which takes 
full observations from different domains (distinguished by the colors in Fig.\ref{fig2}) as input.
Full observations include all the information that an autonomous driving vehicle can observe
in a traffic scene (e.g., history trajectories of neighboring vehicles, map information, etc.),
which are highly domain-related.
$\Phi_v(.)$ is trained using domain contrastive loss~\cite{khosla2020supervised,liu2022towards}.
In order to calculate this loss,
CILF employs dimensional reduction to \textit{V} through a projection layer $p_i=\beta(V_i)$,
where $i$ denotes the domain label.
The contrastive loss between a pair of samples $p_i$ and $p_j$ is defined as follows:
\begin{equation}\label{eq4}
  \mathcal{L}_{contrast}=-\mathrm{log}\frac{\mathrm{exp}(p_i,p_j)/\tau}
  {\Sigma_k\mathbb{I}_{[k=j\lor S_k\neq S_i]}\mathrm{exp}(\mathrm{sim}(p_i,p_k)/\tau)}
\end{equation}
where $\mathbb{I}_{[k=j\lor S_k\neq S_i]}$ denotes the indicator function.
When sample $p_i$ and $p_k$ are from the same domain, this function equals to 1.
$\tau$ is the temperature parameter of contrastive learning,
$\mathrm{sim}(a,b)=\frac{a^{\mathrm{T}}b}{\lVert a\rVert \lVert b\rVert}$
is the cosine similarity.

\subsection{Separate Domain-variant Causal and Non-causal Feature}
In Section 4.1, we argue that domain-invariant features naturally possess causality
as they are the intersection of feature sets from different domains.
However, domain-variant features do not naturally carry
this causal property. It is necessary to filter out the non-causal dimensions
within these features according to causal sufficiency~\cite{peters2017elements,reichenbach1956direction,scholkopf2012causal}:

\begin{definition}
  Causally sufficient feature set: for the prediction task from $X$ to $Y$,
  a feature set is considered causally sufficient if and only if it captures
  all causal effects from $X$ to $Y$.\label{def-2}
\end{definition}

Supervised learning can't guarantee the learned features to be causally sufficient
according to Definition.\ref{def-2}.
Some dimensions may contain more causal information and have a decisive impact on prediction,
while other dimensions may have little influence on prediction.
CILF introduces a neural network based mask generator $M(.)$
that produces a causally sufficient mask~\cite{lv2022causality} using the Gumbel-SoftMax technique~\cite{jang2016categorical}:
\begin{equation}\label{eq5}
  m=\mathrm{Gumbel-SoftMax}(M(V),kN),
\end{equation}
The causally sufficient mask can identify the contribution $k\in (0,1)$ of each dimension
in \textit{V} to prediction. The top $kN$ dimensions with the highest contributions
are regarded as domain-variant causal features \textit{VC},
while the remaining dimensions are considered domain-variant non-causal features \textit{VN}:
\begin{equation}\label{eq6}
  \textit{VC}=V\cdot m,\textit{VN}=V\cdot(1-m).
\end{equation}

\textit{VC} and \textit{VN} are fused with \textit{IC} by the fusion module $F(.)$ to form causal features CF
and spurious features SF, which are then decoded separately by $\Theta(.)$ to 
generate future predictions.
Prediction losses $\mathcal{L}_{causal}$ and
$\mathcal{L}_{spurious}$ can be calculated by RMSE:
\begin{equation}
  \begin{split}\label{eq7}
    CF=F(VC,IC)&=f_2(f_1(VC,IC)+IC),\\
    SF=F(VN,IC)&=f_2(f_1(VN,IC)+IC),
  \end{split}
\end{equation}
\begin{equation}
  \begin{split}\label{eq8}
    \mathcal{L}_{causal}&=\mathrm{RMSE}(\Theta(VC),Y),\\
    \mathcal{L}_{spurious}&=\mathrm{RMSE}(\Theta(VN),Y),
  \end{split}
\end{equation}
where $f_1$ and $f_2$ are FC layers in fusion module.

Decoder $\Theta$ can be trained by minimizing $\mathcal{L}_{causal}$ and
$\mathcal{L}_{spurious}$. Mask generator $M$ can be trained by minimizing
$\mathcal{L}_{causal}$ while adversarially maximizing $\mathcal{L}_{spurious}$.
Overall, the optimization objective can be formulated as follows:
\begin{equation}\label{eq9}
    \underset{\Phi_{IC},F,\Theta}{\mathrm{min}}
    \mathcal{L}_{causal}+\mathcal{L}_{spurious}+\lambda\mathcal{L}_{irm}
\end{equation}
\begin{equation}\label{eq10}
  \underset{\Phi_{v},F,\Theta}{\mathrm{min}}
  \mathcal{L}_{causal}+\mathcal{L}_{spurious}+\alpha\mathcal{L}_{irm}
\end{equation}
\begin{equation}\label{eq11}
  \underset{M}{\mathrm{min}}
  \mathcal{L}_{causal}-\mathcal{L}_{spurious}
\end{equation}
where $\lambda$ and $\alpha$ are balance parameters.

\section{Experiments}
This chapter presents quantitative and qualitative domain generalization experiments
of the proposed CILF framework on the public vehicle trajectory prediction datasets
NGSIM~\cite{punzo2011assessment} and INTERACTION~\cite{zhan2019interaction}.
\subsection{Experiment Design}
INTERACTION is a large-scale real-world dataset that includes top-down
vehicle trajectory data in three scenarios: intersections, highway ramps (referred to as merging),
and roundabouts, collected from multiple locations in America, Asia, and Europe.
INTERACTION consists of 11 subsets, categorized into three types of scenarios:
roundabouts, highway ramps, and intersections (see Table~\ref{tab1}).
NGSIM is a collection of vehicle trajectory data derived from video recordings.
It includes vehicle trajectory data from three different locations:
the US-101 highway, Lankershim Blvd. in Los Angeles, and the I-80 highway in Emeryville.
The sampling rate of both datasets is 10Hz. For each trajectory that lasts 8 seconds, the model takes the first 3 seconds as input
and predicts the trajectory for the next 5 seconds.

As shown in Table~\ref{tab1}, data volume among different subsets
of INTERACTION varies significantly. 
Subsets 0, 1, and 7 have notably higher data volumes compared to other subsets.
To mitigate the potential influence of data volume disparities, we design three contrastive
experiments to demonstrate the effectiveness of CILF in addressing the DG problem:

\begin{table}
  \caption{INTERACTION Dataset. Data Ratio represents the proportion of data volume in each subset
  relative to the total data volume.}\label{tab1}
  \begin{center}
  \renewcommand{\arraystretch}{1}
  \begin{tabular}{ p{2cm}<{\centering}ccc}
  \hline
  Subset ID & Sample Location & Scenario & Data Ratio\\
  \hline
  0 & USA   & Roundabout   & 17.9\% \\
  1 & CHN   & Merging      & 35.2\% \\
  2 & USA   & Intersection &  7.9\% \\
  3 & USA   & Intersection &  2.6\% \\
  4 & GER   & Roundabout   &  1.8\% \\
  5 & USA   & Roundabout   &  5.3\% \\
  6 & GER   & Merging      &  1.2\% \\
  7 & USA   & Intersection & 21.6\% \\
  8 & USA   & Roundabout   &  3.2\% \\
  9 & USA   & Intersection &  2.6\% \\
  10& CHN   & Roundabout   &  0.6\% \\
  \hline
  \end{tabular}
  \end{center}
\end{table}

(1)Single-scenario domain generalization:
Both training and test datasets come from the same scenario within the INTERACTION.
Specifically, the subset with the largest data volume in each scenario is designated as the test set,
while the remaining subsets serve as the training set.

(2)Cross-scenario domain generalization:
The training and test datasets come from different scenarios within the INTERACTION.
Specifically, the three subsets with the largest data volumes (roundabout-0, merging-1, and intersection-7) are
selected as training sets, while the remaining subsets are chosen as test sets.

(3)Cross-dataset domain generalization:
The INTERACTION is selected as the training set, while the NGSIM is selected as the test set.

We choose three classic vehicle trajectory prediction models as baselines:

\textbf{S-LSTM}~\cite{alahi2016social}:An influential method based on a social pooling model;

\textbf{CS-LSTM}~\cite{deo2018convolutional}:A model using a convolutional social pooling structure to learn vehicle interactions;

\textbf{MFP}~\cite{tang2019multiple}:An advanced model that learns semantic latent variables for trajectory prediction.
These baselines are trained under CILF to compare their domain generalization performance.

We employ two commonly used metrics in vehicle trajectory prediction
to measure the model's domain generalization performance:

\textbf{Average Displacement Error (ADE)}: Average L2 distance between predicted points and groundtruth points across all timesteps.
\begin{equation}\label{eq12}
  \mathrm{ADE}=\frac{\Sigma_{i=1}^N\Sigma_{t=t_{obs}+1}^{t_{pred}}\lVert\hat{Y}_{i,t}-Y_{i,t}\rVert}{N},
\end{equation}

\textbf{Final Displacement Error (FDE)}: L2 distance between predicted points and ground truth points at the final timestep.
\begin{equation}\label{eq13}
  \mathrm{FDE}=\frac{\Sigma_{i=1}^N\lVert\hat{Y}_{i,t}-Y_{i,t}\rVert}{N},t=t_{obs}+1,...,t_{pred}.
\end{equation}

\subsection{Quantitative Experiment and Analysis}
\subsubsection{Single-scenario domain generalization}
Table~\ref{tab2} and Table~\ref{tab3} present the comparison
between different models trained under CILF and vanilla conditions
in terms of single-scenario domain generalization on the INTERACTION.
\begin{table}
  \caption{ADE and FDE (in meters) results of domain generalization experiment in the intersection scenario.
  Subset 7 is left for testing, while other subsets of the intersection scenario are used for training.
  }\label{tab2}
  \begin{center}
    \renewcommand{\arraystretch}{1.05}
    \begin{tabular}{|p{2.3cm}<{\centering}|p{1.6cm}<{\centering} p{1.6cm}<{\centering}|
                                           p{1.6cm}<{\centering} p{1.6cm}<{\centering}|
                                           p{1.6cm}<{\centering} p{1.6cm}<{\centering}|}
    \hline
    & \multicolumn{2}{c|}{\bfseries S-LSTM}
    & \multicolumn{2}{|c|}{\bfseries CS-LSTM}
    & \multicolumn{2}{c|}{\bfseries MFP}\\
    & ADE & FDE & ADE & FDE & ADE & FDE\\
    \hline
  
    Intersection-2 &\textbf{1.52}/1.59 & \textbf{4.65}/4.85 &
                    \textbf{1.53}/1.64 & \textbf{4.62}/4.86 &
                    \textbf{1.62}/1.72 & \textbf{5.05}/5.29 \\
    Intersection-3 &\textbf{1.21}/1.29 & \textbf{3.48}/3.60 &
                    \textbf{1.18}/1.28 & \textbf{3.43}/3.59 &   
                    \textbf{1.25}/1.32 & \textbf{3.67}/3.83 \\
    Intersection-9 &\textbf{1.20}/1.29 & \textbf{3.37}/3.53 &
                    \textbf{1.20}/1.30 & \textbf{3.36}/3.56 &
                    \textbf{1.25}/1.34 & \textbf{3.57}/3.79 \\
    \hline
    Intersection-7 &\textbf{2.22}/2.25 & \textbf{6.20}/6.32 &
                    \textbf{2.34}/2.37 & \textbf{6.30}/6.47 &
                    \textbf{2.12}/2.21 & \textbf{6.02}/6.04 \\
    \hline
    \end{tabular}
  \end{center}
  \end{table}
\begin{table}
  \caption{ADE and FDE (in meters) results of domain generalization experiment in the roundabout scenario.
  Subset 0 is left for testing, while other subsets of the roundabout scenario are used for training.
  }\label{tab3}
  \begin{center}
    \renewcommand{\arraystretch}{1.05}
    \begin{tabular}{|p{2.3cm}<{\centering}|p{1.6cm}<{\centering} p{1.6cm}<{\centering}|
                                           p{1.6cm}<{\centering} p{1.6cm}<{\centering}|
                                           p{1.6cm}<{\centering} p{1.6cm}<{\centering}|}
    \hline
    & \multicolumn{2}{c|}{\bfseries S-LSTM}
    & \multicolumn{2}{|c|}{\bfseries CS-LSTM}
    & \multicolumn{2}{c|}{\bfseries MFP}\\
    & ADE & FDE & ADE & FDE & ADE & FDE\\
    \hline
  
    Roundabout-4 &\textbf{1.68}/1.75 & \textbf{4.41}/4.64 &
                  \textbf{1.62}/1.83 & \textbf{4.41}/4.81 &
                  \textbf{1.59}/1.73 & \textbf{2.86}/3.14 \\
    Roundabout-6 &\textbf{1.03}/1.12 & \textbf{2.97}/3.09 &
                  \textbf{1.01}/1.10 & \textbf{2.86}/3.02 &
                  \textbf{1.01}/1.14 & \textbf{2.86}/3.14 \\
    Roundabout-10&\textbf{1.42}/1.57 & \textbf{3.60}/3.94 &
                  \textbf{1.38}/1.51 & \textbf{3.50}/3.94 &
                  \textbf{1.30}/1.60 & \textbf{3.42}/4.05 \\
    \hline
    Roundabout-0 &\textbf{3.31}/3.34 & \textbf{9.04}/9.38 &
                  \textbf{3.24}/3.35 & \textbf{9.01}/9.09 &
                  \textbf{3.15}/3.25 & \textbf{8.80}/8.83 \\
    \hline
    \end{tabular}
  \end{center}
  \end{table}
CILF achieves both ADE and FDE improvements
for all models in both the source and target domains.

In the intersection scenario, as shown in Table~\ref{tab2},
for the source domains, CS-LSTM achieves the largest improvement under CILF,
with average increments of 7.40\% and 5.00\% (ADE and FDE).
Conversely, S-LSTM shows the lowest improvement,
with average increments of 5.86\% and 4.00\%.
As for the target domain, S-LSTM and MFP demonstrate similar improvements,
with increments of 1.33\%, 1.90\% for S-LSTM and 1.40\%, 1.33\% for MFP.

In the roundabout scenario, as shown in Table~\ref{tab3},
for the source domains, MFP achieves the largest improvement under CILF,
with average increments of 12.75\% and 9.31\% (ADE and FDE).
Conversely, S-LSTM shows the lowest improvement,
with average increments of 7.20\% and 5.82\%.
As for the target domain, S-LSTM demonstrates the largest improvements,
with increments of 3.90\% and 3.62\%.
MFP achieves the lowest improvement,
with increments of 3.08\% and 2.34\%.
Obviously, in the roundabout scenario, CILF exhibits larger improvements
in both the source and target domains compared to intersection.

\subsubsection{Cross-scenario domain generalization}
Table~\ref{tab4} presents the comparison
between different models trained under CILF and vanilla conditions
in terms of cross-scenario domain generalization on the INTERACTION.
CILF achieves both ADE and FDE improvements
for all models in both the source and target domains.
For the source domains, CS-LSTM achieves the largest improvement under CILF,
with average increments of 7.54\% and 5.95\% (ADE and FDE).
Conversely, MFP shows the lowest improvement,
with average increments of 3.58\% and 2.95\%.
As for the target domain, both three models achieve similar improvements,
with increments of 4.73\%, 3.53\% for S-LSTM;
5.47\%, 3.17\% for CS-LSTM; and 3.65\%, 2.53\% for MFP.

\begin{table}
  \caption{ADE and FDE (in meters) results of cross-scenario domain generalization experiment.
  Subset 0, 1, and 7 are left for training, while other subsets of INTERACTION are used for testing.
  }\label{tab4}
  \begin{center}
    \renewcommand{\arraystretch}{1.05}
    \begin{tabular}{|p{2.3cm}<{\centering}|p{1.75cm}<{\centering} p{1.75cm}<{\centering}|
                                            p{1.75cm}<{\centering} p{1.75cm}<{\centering}|
                                            p{1.75cm}<{\centering} p{1.75cm}<{\centering}|}
    \hline
    & \multicolumn{2}{c|}{\bfseries S-LSTM}
    & \multicolumn{2}{|c|}{\bfseries CS-LSTM}
    & \multicolumn{2}{c|}{\bfseries MFP}\\
    & ADE & FDE & ADE & FDE & ADE & FDE\\
    \hline
  
    Roundabout-0  &\textbf{1.40}/1.51 & \textbf{4.33}/4.58 &
                   \textbf{1.36}/1.46 & \textbf{4.24}/4.46 &
                   \textbf{1.45}/1.50 & \textbf{4.47}/4.56 \\
    Merging-1     &\textbf{0.83}/0.85 & \textbf{2.28}/2.33 &
                   \textbf{0.80}/0.86 & \textbf{2.24}/2.36 &
                   \textbf{0.86}/0.89 & \textbf{2.48}/2.55 \\
    Intersection-7&\textbf{1.18}/1.26 & \textbf{3.60}/3.78 &
                   \textbf{1.14}/1.25 & \textbf{3.53}/3.83 &
                   \textbf{1.19}/1.24 & \textbf{3.70}/3.86 \\
    \hline
    Intersection-2&\textbf{2.42}/2.52 & \textbf{7.59}/7.82 &
                   \textbf{2.45}/2.55 & \textbf{7.72}/7.90 &
                   \textbf{2.42}/2.48 & \textbf{7.60}/7.72 \\
    Intersection-3&\textbf{1.63}/1.73 & \textbf{4.92}/5.23 &
                   \textbf{1.68}/1.77 & \textbf{5.13}/5.42 &
                   \textbf{1.66}/1.71 & \textbf{5.02}/5.18 \\
    Roundabout-4  &\textbf{3.32}/3.67 & \textbf{10.13}11.01 &
                   \textbf{3.42}/3.65 & \textbf{10.60}/10.76 &
                   \textbf{3.59}/3.78 & \textbf{10.71}/11.26 \\
    Roundabout-5  &\textbf{1.75}/1.83 & \textbf{5.16}/5.38 &
                   \textbf{1.72}/1.84 & \textbf{5.20}/5.36 &
                   \textbf{1.71}/1.82 & \textbf{4.97}/5.25 \\
    Merging-6     &\textbf{1.19}/1.21 & \textbf{3.59}/3.63 &
                   \textbf{1.15}/1.24 & \textbf{3.54}/3.69 &
                   \textbf{1.07}/1.14 & \textbf{3.34}/3.43 \\
    Roundabout-8  &\textbf{2.07}/2.14 & \textbf{6.21}/6.34 &
                   \textbf{2.07}/2.22 & \textbf{6.21}/6.48 &
                   \textbf{2.08}/2.13 & \textbf{6.17}/6.28 \\
    Intersection-9&\textbf{1.65}/1.70 & \textbf{4.90}/5.04 &
                   \textbf{1.66}/1,75 & \textbf{4.99}/5.19 &
                   \textbf{1.62}/1.64 & \textbf{4.86}/4.86 \\
    Roundabout-10 &\textbf{2.80}/2.99 & \textbf{8.64}/8.76 &
                   \textbf{2.84}/2.92 & \textbf{8.85}/8.75 &
                   \textbf{2.86}/2.95 & \textbf{8.50}/8.59 \\
    \hline
    \end{tabular}
  \end{center}
\end{table}

\subsubsection{Cross-dataset domain generalization}
Table \ref{tab5} presents the comparison
between different models trained under CILF and vanilla conditions
in terms of cross-dataset domain generalization.
CILF achieves both ADE and FDE improvements
for all models in both the source and target domains.
For the source domains, CS-LSTM achieves the largest improvement under CILF,
with average increments of 7.54\% and 5.95\% (ADE and FDE).
Conversely, MFP shows the lowest improvement,
with average increments of 3.58\% and 2.95\%.
For the target domains, S-LSTM achieves the largest improvement under CILF,
with average increments of 4.39\% and 3.53\%.
Conversely, MFP shows the lowest improvement,
with average increments of 1.12\% and 0.64\%.
\begin{table}
  \caption{ADE and FDE (in meters) results of cross-dataset domain generalization experiment.
  Subset 0, 1, and 7 of the INTERACTION are left for training,
  while the NGSIM is used for testing.}\label{tab5}
  \begin{center}
    \renewcommand{\arraystretch}{1.05}
    \begin{tabular}{|p{2.3cm}<{\centering}|p{1.6cm}<{\centering} p{1.6cm}<{\centering}|
                                            p{1.6cm}<{\centering} p{1.6cm}<{\centering}|
                                            p{1.6cm}<{\centering} p{1.6cm}<{\centering}|}
    \hline
    & \multicolumn{2}{c|}{\bfseries S-LSTM}
    & \multicolumn{2}{|c|}{\bfseries CS-LSTM}
    & \multicolumn{2}{c|}{\bfseries MFP}\\
    & ADE & FDE & ADE & FDE & ADE & FDE\\
    \hline
    Roundabout-0  &\textbf{1.40}/1.51 & \textbf{4.33}/4.58 &
                   \textbf{1.36}/1.46 & \textbf{4.24}/4.46 &
                   \textbf{1.45}/1.50 & \textbf{4.47}/4.56 \\
    Merging-1     &\textbf{0.83}/0.85 & \textbf{2.28}/2.33 &
                   \textbf{0.80}/0.86 & \textbf{2.24}/2.36 &
                   \textbf{0.86}/0.89 & \textbf{2.48}/2.55 \\
    Intersection-7&\textbf{1.18}/1.26 & \textbf{3.60}/3.78 &
                   \textbf{1.14}/1.25 & \textbf{3.53}/3.83 &
                   \textbf{1.19}/1.24 & \textbf{3.70}/3.86 \\
    \hline
    NGSIM         &\textbf{3.27}/3.42 & \textbf{8.47}/8.78 &
                   \textbf{3.50}/3.60 & \textbf{9.17}/9.47 &
                   \textbf{3.54}/3.58 & \textbf{9.28}/9.38 \\
    \hline
    \end{tabular}
  \end{center}
\end{table}

\subsection{Qualitative Experiment and Analysis}
This section presents the comparison of predicted trajectories generated by
CILF-MFP and Vanilla-MFP in the cross-scenario domain generalization experiment.

\begin{figure}
  \centering
  \subfigure[CILF-MFP]{
    \includegraphics[width=0.45\linewidth]{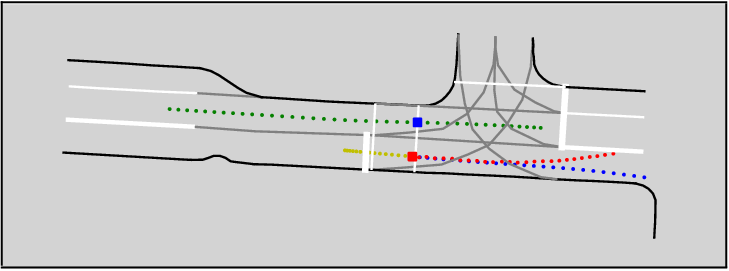}
    \label{fig3-1}
  }
  \subfigure[Vanilla-MFP]{
    \includegraphics[width=0.45\linewidth]{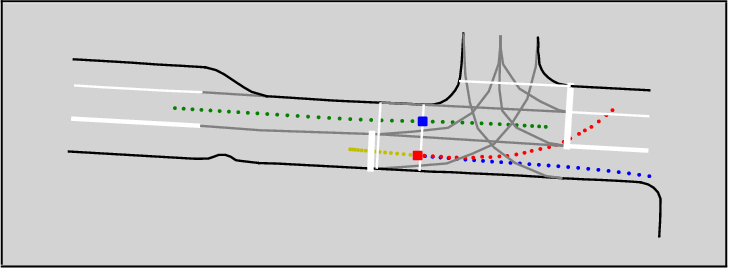}
    \label{fig3-2}
  }
  \centering
  \subfigure[CILF-MFP]{
    \includegraphics[width=0.45\linewidth]{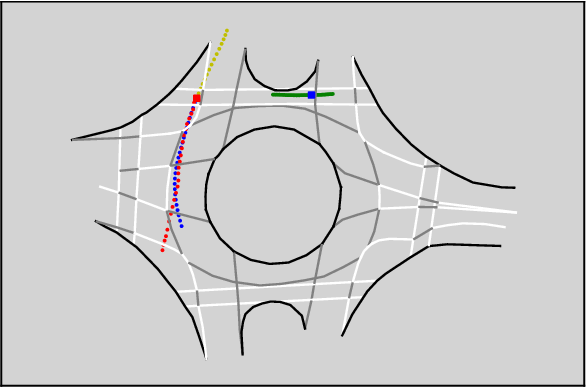}
    \label{fig3-3}
  }
  \subfigure[Vanilla-MFP]{
    \includegraphics[width=0.45\linewidth]{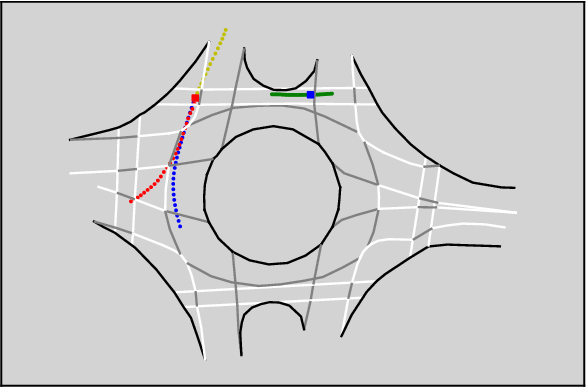}
    \label{fig3-4}
  }
  \caption{
    Contrastive experiments between CILF-MFP and Vanilla-MFP on two target domains.
    (a) and (b) demonstrate the contrastive experiment on subset-3 (intersection) of INTERACTION,
    (c) and (d) demonstrate the contrastive experiment on subset-4 (roundabout) of INTERACTION.
  }\label{fig3}
\end{figure}

Fig.\ref{fig3} illustrates the comparison of predicted trajectories between
CILF-MFP and MFP in target domain subset-3 (intersection) and subset-4 (roundabout).
The black bold solid lines represent the curb,
and the white and gray bold solid lines represent guide lines.
The red square denotes the current vehicle,
and the blue square represents neighboring vehicles perceived by the current vehicle.
The yellow dots represent history trajectories of the current vehicle,
while the blue dots represent future trajectories.
The red dots represent the predicted future trajectory. 
The green dots represent the historical and future trajectories of neighboring vehicles.

In both scenarios, CILF-MFP demonstrates remarkable improvement
in prediction quality compared to MFP,
particularly at the end of the predicted trajectories.

\section{Conclusion}
To improve the generalization ability of vehicle trajectory prediction models,
we first analyze the causal structure of OOD generalization and propose OOD-CG,
which highlights the limitations of conventional correlation-based learning framework.
Then we propose CILF to employ only causal features for prediction by three steps:
(a) extracting \textit{IC} by invariant risk minimization,
(b) extracting \textit{V} by domain contrastive learning, and
(c) separating \textit{VC} and \textit{VN} by domain adversarial learning.
Quantitative and qualitative experiments on several mainstream
datasets prove the effectiveness of our model.

\bibliographystyle{splncs04}
\bibliography{./ref}
\end{document}